\documentclass[letterpaper, 10 pt, conference]{ieeeconf}  % Comment this line out if you need a4paper

\IEEEoverridecommandlockouts                              % This command is only needed if 
\usepackage{multicol}
\usepackage[bookmarks=true]{hyperref}

\usepackage{amsmath}
\usepackage{amssymb}
\usepackage{graphicx}
\usepackage{caption}
\usepackage{subcaption}
\usepackage{url}
\usepackage{colortbl}

\usepackage{algorithm}
\usepackage[noend]{algpseudocode}
\DeclareMathOperator*{\argmin}{argmin} 
\DeclareMathOperator*{\argmax}{argmax}

\title{\LARGE \bf
	Learning Implicit Sampling Distributions for Motion Planning
}

\author{Clark Zhang, Jinwook Huh, and Daniel D. Lee\\% <-this % stops a space
	\thanks{All authors are affiliated with the GRASP Laboratory, University of Pennsylvania, Philadelphia, PA 19104. {\tt\small\{clarkz,jinwookh, ddlee\}@seas.upenn.edu}}
}

\begin{document}
	\maketitle
	\thispagestyle{empty}
	\pagestyle{empty}
	
	\begin{abstract}
		Sampling-based motion planners have experienced much success due to their ability to efficiently and evenly explore the state space. However, for many tasks, it may be more efficient to not uniformly explore the state space, especially when there is prior information about its structure.   Previous methods have attempted to modify the sampling distribution using hand selected heuristics that can work well for specific environments but not universally. In this paper, a policy-search based method is presented as an adaptive way to learn implicit sampling distributions for different environments. It utilizes information from past searches in similar environments to generate better distributions in novel environments, thus reducing overall computational cost. Our method can be incorporated with a variety of sampling-based planners to improve performance. Our approach is validated on a number of tasks, including a 7DOF robot arm, showing marked improvement in number of collision checks as well as number of nodes expanded compared with baseline methods.
	\end{abstract}
	
	\IEEEpeerreviewmaketitle
	
	%%%%%%%%%%%%%%%%%%%%%%%%%%%%%%%%%%%%%%%%%%%%%%%%%%%%%%%%%%%%%%%%%%%%%%%%%%%%%%%%
	\section{INTRODUCTION}
	Sampling-based motion planners are efficient tools to plan in high dimensional spaces in difficult environments. They can be used to plan motions for robotic manipulation tasks, autonomous car maneuvers, and many other problems. An important aspect of a sampling-based motion planner is the sampling distribution. Planners such as Probabilistic Road Map (PRM) \cite{Kavraki1996}, Rapidly-Exploring Random Trees (RRT) \cite{Lavalle2001}, Expansive Space Trees (EST) \cite{Hsu1997}, Fast Marching Trees (FMT*) \cite{Janson2015}, Batch Informed Trees (BIT*) \cite{Gammell2015}, etc. and their many variants iteratively build trees to connect samples drawn from their sampling distributions. Thus, the distribution strongly affects how the search progresses. Traditionally, planners draw random state samples from a uniform distribution (many times with a slight goal bias). However, for many classes of environments, a different probability distribution over the state space can speed up planning times. For example in environments with sparse obstacles, it can be useful to heavily bias the samples towards the goal region as the path to the goal will be relatively straight. The natural questions to ask are ``How heavily should the goal be biased?" or more generally ``What is the best probability distribution to draw out of?" In previous literature, many researchers have found good heuristics \cite{Yershova2005} \cite{Shkolnik2011} to modify the probability distributions for specific environments. However, these heuristics do not work generally and may not apply to new environment types. In fact, a heuristic can increase the planning time dramatically if it is unsuited to the problem at hand.
	
	In this work, we present a systematic way to generate effective probability distributions automatically for different types of environments. The first issue encountered is how to choose a good representation for probability distributions. The sampling distributions can be very complicated in shape and may not easily be representable as common distributions such as Gaussians or Mixtures of Gaussians. Instead, the distribution is represented with rejection sampling, a powerful method that can implicitly model intricate distributions. The process of accepting or rejecting samples is formulated as a Markov Decision Process. This way, policy gradient methods from traditional Reinforcement Learning literature can be used to optimize the sampling distribution for any planning costs such as the number of collision checks or the planning tree size. The method presented will use past searches in similar environments to learn the characteristics of good planning distributions. Then, the rejection sampling model will be applied to new instances of the environment. 
	
	\begin{figure}
		\centering
		\includegraphics[width=\columnwidth]{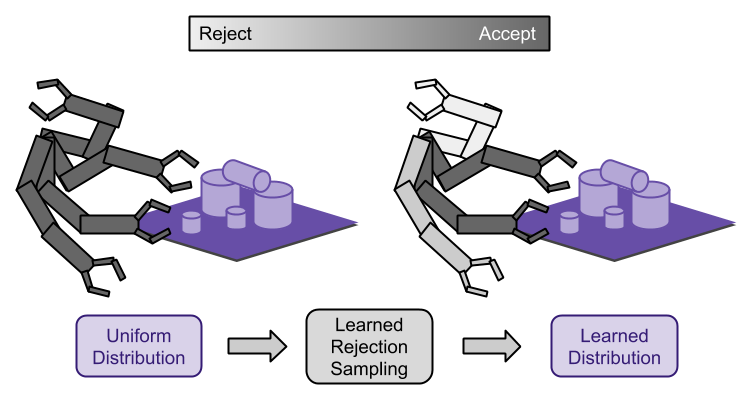}
		\caption{An example of a learned distribution for the task of a robot arm reaching for various objects on a tabletop. On the left, samples from a uniform distribution of the configuration space are displayed. Some of which may be rejected by a learned rejection sampling policy to form a new learned distribution over the configuration space.}
		\label{arm_distribution}
	\end{figure}

	The contribution of this paper is to 1) present an adaptive approach to generating good probability distributions for different environments that improve the performance of sampling based planners and 2) analyze the policies learned against previous heuristic approaches. The method presented is shown to imitate previous heuristic approaches on a simple 2D problem and has found good intuitive heuristics for tabletop manipulation tasks for robotic arms. The paper is organized as follows: Section \ref{RelatedWork} discusses previous research in modifying sampling distributions. Section \ref{ProblemStatement} gives a formal view of the problem. Section \ref{LearningSamplingDistributions} describes our method. Section \ref{ImplementationDetails} gives specific implementation details. Section \ref{Experiments} details an experiment on a simulated environment as well as a real robot experiment and conclusions are presented in Section \ref{Conclusion}.

	\section{Related Work}\label{RelatedWork}
	There is a number of methods that use rejection sampling to bias the sampling distribution. Boor \textit{et al.} \cite{Boor1999} introduces a method to bias random samples towards obstacles for the PRM planner. For every sampled state, an addition state is generated from a Gaussian distribution around the first state. A sample is only accepted if exactly one point is in collision. Urmson and Simmons \cite{Urmson2003} proposed a method to compute lower cost RRT paths. Each node in their tree is given a heuristic ``quality" that estimates how good a path passing through that node will be. Rejection sampling is used to sample points near high quality nodes. This method is mostly superseded by RRT* \cite{Karaman2010}, but is a useful case of how rejection sampling has been used to improve path quality. Yershova \textit{et al.} \cite{Yershova2005} introduces Dynamic-Domain RRT which rejects samples that are too far from the tree. The idea is that drawing samples on the other side of an obstacle is wasteful since it will lead to a collision, so sampling is restricted to an area close to the tree. Shkolnik and Tedrake \cite{Shkolnik2011} introduce BallTree which does the opposite of Dynamic-Domain RRT and rejects samples that are too close to the tree. The idea is that many nodes in the tree are wasted in exploring areas that are close. Shkolnik and Tedrake \cite{Shkolnik2009-2} also present another heuristic to improve RRT performance for differentially constrained systems by rejecting samples where the reachability region of the nearest neighbor is further from the random sample than the nearest neighbor itself, so that extending towards the sample will not actually encourage exploration. 
	
	There are also methods that do not utilize rejection sampling. Zhang and Manocha \cite{Zhang2008} modifies random samples by moving points in the obstacle space to the nearest point in free space. The effect of this method is that small ``tunnels" that are surrounded by obstacles will be sampled more frequently. As they have noted, this is effective for environments that have narrow passages which are particularly hard for traditional planners to solve due to the small probability of sampling within the narrow passage. Diankov \textit{et al.} \cite{Diankov2008} grows a backward tree in the task space and biases samples in the forward configuration space tree towards it. The backward task space tree can be much more easily found in manipulation tasks and can effectively guide the forward configuration space tree. Yang and Brock \cite{Yang2004} propose a method to quickly compute an approximation to the medial axis of a workspace. Their goal is to generate PRM samples that are close to the medial axis, as it is a good heuristic to plan in environments with narrow tunnels. This has also been explored in \cite{wilmarth1999maprm}.
	
	While the previous work has yielded good results for certain environments, they are not generally applicable. There has been some work in automating how to improve sampling for different environments. Zucker \textit{et al.} \cite{Zucker2008} introduces a method to optimize workspace sampling. The workspace is discretized and features such as visibility are computed for each discrete cell. The workspace sampling is improved using the REINFORCE algorithm \cite{Williams1992}. This method performs well in the environment it is optimized in, but new environments can potentially have a high preprocessing cost to compute the features. In addition, discretizing the workspace may be infeasible for certain problem domains. Gammell \textit{et al.} \cite{Gammell2014} introduced Informed RRT* which improves RRT* performance by restricting samples to an ellipsoid that contains all samples that could possible improve the path length after an initial path is found. Kunz \textit{et al.} \cite{Kunz2016} and Yi \textit{et al.} \cite{Yi2017} improve the informed sampling technique. This technique does not improve the speed at which the first path is found. More recently, Ichter \textit{et al.} \cite{Ichter2017} used a Variational Autoencoder to learn an explicit sampling distribution for FMT*. 
	
	Our approach differs from previous work by introducing a general method for sampling based planners that is not a human created heuristic nor does it require any discretization of the workspace. In addition, this method can be combined with most previous approaches to further improve performance.

	\section{Problem Statement}\label{ProblemStatement}
	The problem this paper addresses is to reduce the computational cost of path planning in certain types of environments by modifying the sampling distributions. For clarity, let us consider planning trajectories for a robotic arm in typical tabletop environments.
	
	Following the notation from \cite{Karaman2010}, a state space for a planning problem is denoted as $X$. For a given environment, let $X_{obs}$ denote the obstacle space, a subset of $X$ that the robot can not move in. Thus a map is uniquely defined by its $X_{obs}$. A specific environment type, $E$, is a probability distribution over possible obstacle spaces, $X_{obs}$. For a 7DOF robotic arm, $X$ is the 7 dimensional configuration space, and $E$ will assign higher probability to environments that look like scattered objects on a table. 
   
   Let $\boldsymbol{Y}_k \sim \mu_k$ be a sequence of Random Variables that represents the $i^{th}$ random sample of the state space drawn during the planning process (Note that the random variables do not need to be identical, as shown in Fig. \ref{Adistributions}). In standard sampling-based planners, $\boldsymbol{Y}_k$ are independent and identically distributed. 
	Now given a specific map, $X_{obs}$, and a sequence of random state space samples, let $\boldsymbol{Z}(X_{obs},  Y_1, Y_2, \hdots)$ be a Random Variable representing the number of collision checks, the size of the search tree, and the number of random samples drawn during the planning process. $\boldsymbol{Z}$ is a Random Variable due to its dependence on the random samples, $\boldsymbol{Y}_i$, that are drawn.  The problem this paper addresses is the following optimization problem:
	\begin{eqnarray}\label{opt1}
	\{\mu_k^*\} = \argmin_{\{\mu_k\}} \mathbb{E}_{X_{obs}, \{\mu_k\}} [ \boldsymbol{Z}(X_{obs}, Y_1, Y_2, \hdots)].
	\end{eqnarray}
	
	Eq. \ref{opt1} succinctly describes the following: Given a distribution of maps, $E$, find the sequence of distributions $\{\mu_k^*\}$ that minimizes the expected computational cost of the search, $\boldsymbol{Z}$. For a robotic arm, this amounts to finding the probability distribution that will minimize the number of collision checks, size of the search tree, and the number of random samples drawn in common tabletop environments.

	\section{Learning Sampling Distributions} \label{LearningSamplingDistributions}
	It is difficult to represent the sequence of distributions, $\mu_k$ from Eq. \ref{opt1} explicitly. The distribution may be very complicated and not easily representable with simple distributions. In addition, there may not be an easy explicit map available (often there is just an oracle that returns whether a collision has occurred). A way to implicitly represent a complicated distribution is with rejection sampling, similar to techniques presented in \cite{Boor1999}, \cite{Yershova2005}, \cite{Shkolnik2011}, \cite{Shkolnik2009-2}. In our method, random samples will be drawn from some explicitly given distribution, $\nu$ (usually the uniform distribution with a peak at the goal). For each random sample $x \in X$ drawn, a probability of rejection is computed. The sample is then either passed to the planner or rejected. The end result is that unfavorable samples are discarded with high probability so computation time is not wasted in attempting to add the node into the tree or in checking it for collisions. This can improve performance as the sampling operation is usually cheap, but collision checking and tree extension is much more expensive. For example, in the robotic arm experiments described later, the policy has learned that samples with large distances between joints and obstacles are unfavorable as it does not progress the search. The policy is learned offline, and is applied to new environments that are similar in nature (for example, in a grasping task, a desk with different objects in different locations).
	
	More formally, the probability of rejecting a sample $x \in X$ is denoted as $\pi(a_{reject} | x)$ where $a_{reject}$ is the action of rejecting a sample. The function, $\pi$ is learned offline (discussed in Section \ref{solveMDP}). Thus, $\pi$ can implicitly represent a probability measure $\mu$, the distribution that is effectively being sampled when applying rejection sampling.
	\begin{equation}\label{rejectionsampling}
	\mu(X_S \subset X) = 
	\frac{\int_{X_S}(1 - \pi(a_{reject} | x)) \mathrm{d}\nu(x)}
	{\int_{X}(1 - \pi(a_{reject} | x)) \mathrm{d}\nu(x)}
	\end{equation}
	This $\mu$ is valid as long as $\int_{X}(1 - \pi(a_{reject} | x)) \mathrm{d}\nu(x)$ is some finite positive number. This will be easily satisfied if $\pi(a_{reject} | x) < 1, \forall x \in X$.
	
	\subsection{Rejection Sampling as a Markov Decision Process}
	
	\begin{figure}
		\centering
		\includegraphics[width=0.7\columnwidth]{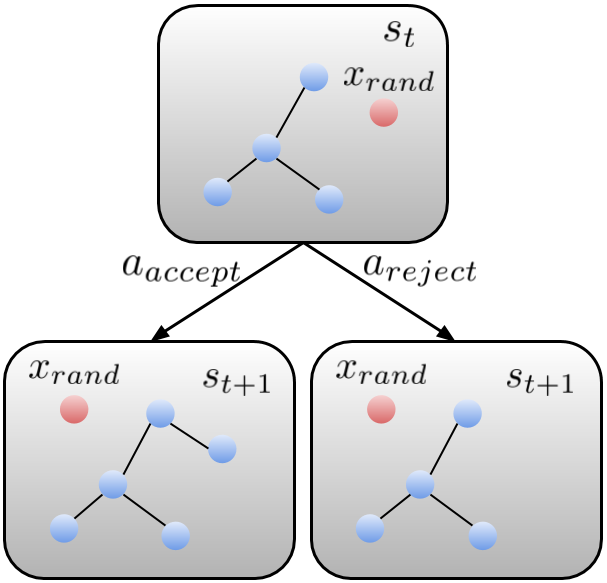}
		\caption{MDP representing rejection sampling in a RRT. Blue circles represent nodes in the tree, while the lines represent edges connecting nodes. At a state $s_t$, you can transition to possible next states, $s_{t+1}$, by rejecting or accepting the random sample $x_{rand}$.}
		\label{MDP}
	\end{figure}

	The process of rejecting samples during the planning algorithm will be modeled as a Markov Decision Process (MDP), so that traditional reinforcement learning methods may be easily applied to optimize the rejection sampling. Following the notation in \cite{Sutton2000}, a MDP consists of a tuple containing $(S, A, \mathcal{P}, r)$ (in some literature, the MDP also includes a discount factor, $\gamma$). $S$ is the set of all possible states the system can be in. $A$ is the set of actions that can be taken. $\mathcal{P}^{a}_{s, s'} = Pr\{s_{t+1}=s' | s_t=s, a_t = a\}$ are transition probabilities, and $r(s, a)$ is the reward for taking action $a \in A$ in state $s \in S$. A typical goal in a MDP is to select a policy, $\pi$, mapping states to actions that maximizes the sum of rewards over a time horizon $T$, $J = \mathbb{E} [ \sum_{t=0}^{T} r(s_t, a_t) ]$.
	
	In the setting of sampling-based planners, a state $s_t \in S$ consists of the environment, $X_{obs}$, the current state of the planner, and a randomly generated random sample $x_t \in X$ from the distribution $\nu$. The action space is $A = \{a_{accept}, a_{reject}\}$. Upon taking action $a_{accept}$, the sample $x_t \in X$ will be passed to the planner. Upon taking action $a_{reject}$, the sample will be rejected. In both cases, a new random sample, $x_{t+1} \in X$ will be included in the new state $s_{t+1}$. A reward, $r(s_t, a_t)$ is given based on $\boldsymbol{Z}(X_{obs}, Y_1, Y_2, \hdots)$. The cost defined for $Z$ will simply become the negative reward. A MDP model of the rejection sampling applied to RRT is described pictorially in Fig. \ref{MDP}. Note that algorithms that may use batches of samples such as PRM or BIT* can utilize this simply by drawing and rejecting samples until there is enough for a batch.
	
	The policy will be defined as $\pi(a | s)$, the probability of taking action $a$ in state $s$. Furthermore, $\pi$ will be restricted to a class of functions with parameters $\theta$ and take in as input a feature vector $\phi(s)$ instead of the raw state $s$. The policy  will be referred to as $\pi_\theta(a | \phi(s))$. In this paper, the function is represented as a neural net where $\theta$ represents the weights in the network.
	By implicitly defining probabilities $\mu_k$ in Eq. \ref{rejectionsampling} with policy $\pi_\theta$, $\mu_k$ can be written as a function of $\theta$. The optimization problem in Eq. \ref{opt1} can be rewritten as
	\begin{eqnarray}\label{opt2}
	\theta^* = \argmin_{\theta} \mathbb{E}_{X_{obs}} [ \boldsymbol{Z}(X_{obs}, Y_1(\theta), Y_2(\theta), \hdots)].
	\end{eqnarray}
	where all $\mu_k$ share the same parameters $\theta$ but may be different distributions due to the different states the planner will be in.
	Furthermore, to keep notation with the reinforcement learning literature, the planning cost, $\boldsymbol{Z}$, will be redefined as
	\begin{equation}
	\boldsymbol{Z}(X_{obs}, P, Y_1, Y_2, \hdots) = -\sum_{t=0}^{T} r_{X_{obs}}(s_t, a_t)
	\end{equation}
	where the rewards $r_{X_{obs}}(s_t, a_t)$ have been chosen to reflect the negative cost represented by $\boldsymbol{Z}(X_{obs}, Y_1, Y_2, \hdots)$. Specific reward functions for experiments are described in Section \ref{ImplementationDetails}. Finally, the expectation can be approximated with some samples of typical environments that $E$ contains.
	\begin{eqnarray}\label{opt3}
	\theta^* = \argmax_{\theta} \frac{1}{|I_E|} 
	\sum_{X_{obs} \in I_E} \mathbb{E}_{\{a_i\} \sim \pi_\theta}[r_{X_{obs}}(s_t, a_t)].
	\end{eqnarray}
	where $I_E$ is a set of $X_{obs}$ that are representative of the environment $E$.
	
	\subsection{Optimizing the Probability Distributions}\label{solveMDP}
	
	There are many methods from reinforcement learning literature that has been developed to solve the optimization problem posed in Eq. \ref{opt3}. These methods can be roughly split into two categories: 1) value based methods such as Q-Learning \cite{Watkins1992} which try to estimate the expected sum of rewards at a given state and 2) policy gradient methods which attempt to directly optimize the policy. This paper utilizes policy gradient methods, in particular, the REINFORCE algorithm introduced by Williams \cite{Williams1992} and later extended to function approximations by Sutton \textit{et al.} \cite{Sutton2000}. The rationale for choosing policy gradient methods over value based methods is that the policy will have an explicit form that is fast to evaluate which is vital as the policy will be used in the innerloop of sampling-based planners.
	
	In REINFORCE with function approximations, the policy $\pi_\theta$ is improved iteratively by taking gradient ascent steps, $\nabla J_{\theta}$, where $J$ is $
	\mathbb{E} [\sum_{t=0}^{T} r_{X_{obs}}(s_t, a_t)]$, the quantity being maximized in Eq. \ref{opt3}. For multiple environments, this can be achieved by iteratively take gradient descent steps for every environment or use an average gradient of all the environments. The likelihood ratio policy gradient presented in \cite{Sutton2000} can be written as
	\begin{equation}\label{policygradient}
	\nabla J_{\theta} =
	\mathbb{E}[\sum_{t=0}^{T}\nabla log(\pi_\theta(a_t| \phi(s_t))) (R_t^{X_{obs}} - V(\phi(s_t)))]
	\end{equation}
	where $R_t^{X_{obs}} = \sum_{k=t}^{T} r_{X_{obs}}(s_k, a_k)$ and $V(\phi(s_t))$ is an estimate of the value function used as a baseline to reduce variance \cite{Sutton2000}. 
	Given an environment $X_{obs}$ and policy $\pi_\theta$, the expectation in Eq. \ref{policygradient} can be estimated by running the planner $N$ times with $\pi_\theta$ and collecting samples of $(\phi(s_t), a_t, r_{X_{obs}}(s_t, a_t), V(\phi(s_t)))$ tuples to calculate $\sum_{t=0}^{T}\nabla log(\pi_\theta(a_t|\phi(s_t))) (R_t^{X_{obs}} - V(\phi(s_t)))$ for each rollout, then averaging over the $N$ rollouts.
	
	During training, another neural network is fitted to represent $V_{w}(\phi(s_t))$ with weights $w$. Utilizing the samples $(\phi(s_t), a_t, r_{X_{obs}}(s_t, a_t), V(\phi(s_t)))$ in each iteration of the policy gradient ascent, an iteration of gradient descent is run on $w$ to minimize the loss function
	\begin{equation}\label{valueloss}
	L = \sum_{t=0}^{T} (V(\phi(s_t)) - R^{X_{obs}}_t)^2.
	\end{equation}
	to update the baseline $V(\phi(s_t))$.
	The steps of the algorithm are detailed in Algorithm \ref{learnalgo}.

	One downside of policy gradient methods is that they are susceptible to local minima as the objective function is not convex. To mitigate this, several different policies are initialized and the best policy is chosen. Different features should also be tested. The performance depends on what information is available.

	\begin{algorithm}
		\caption{Learning Sample Distribution}\label{learnalgo}
		\begin{algorithmic}[1]
			\Procedure{Learn}{$\{X_{obs}^{(i)}\}_{i=1}^{M}$}
			\State Initialize parameters $\theta_0$ for policy $\pi_{\theta_0}$
			\State Initialize parameters $w_0$ for value baseline, $V_{w_0}$
			\State Run planner with $\pi_{\theta_0}$ several times with each environment and collect data $D_0 = (\phi(s_t), a_t, r(s_t, a_t), V_{w_0}(\phi(s_t)))$
			\State Use $D_0$ to fit $V_{w_0}$ by running gradient descent on the loss function in Eq. \ref{valueloss}
			\For{i=1:NumIterations}
			\For {each environment in $I_E$}
			\State Run $\pi_{\theta_{i-1}}$ N times and collect data $D_{i, j}$
			\State Use Eq. \ref{policygradient} to compute gradient and obtain $\pi_{\theta_i}$
			\State Compute gradient of Eq. \ref{valueloss} to obtain $w_{i}$    
			\EndFor
			\EndFor
			\EndProcedure
		\end{algorithmic}
	\end{algorithm}

	\subsection{Probabilistic Completeness}\label{ProbabilisticCompleteness}
	It is intuitive that this process of rejection sampling will preserve probabilistic completeness for RRT. Following the original proof in \cite{Lavalle2001}, the existence of an attraction sequence of length $K$ between the start and goal positions is assumed. The proof then turns into showing that there is a minimum probability of transitioning from one ball in the attraction sequence to the next. Treating the transition as a biased coinflip with success rate $p$, the question of whether a path is found in $N$ steps turns into a question of whether or not out of $N$ coinflips, $K$ are successful. 
	In \cite{Lavalle2001}, $p$ is given as
	\begin{equation}
	p = \min_i \{\nu(A_i) / \nu(X_{free}) \}
	\end{equation}
	where $A_i$ is the $i^{th}$ element in the attraction sequence. The rejection sampling modifies $\nu(A_i)$ and not $\nu(X_{free})$. Setting a lower threshold for the probability of acceptance of a sample as $\epsilon$, we can write
	\begin{align}
	\mu_k(A_i)
	&=  \\
	&\frac{\int_{A_i} \pi(a_{accept} | x) \mathrm{d}\nu(x)}
	{\int_{A_i}\pi(a_{accept} | x) \mathrm{d}\nu(x) + \int_{X \setminus A_i}\pi(a_{accept} | x) \mathrm{d}\nu(x)}\\
	&\geq
	\frac{\int_{A_i} \epsilon \mathrm{d}\nu(x)}
	{\int_{A_i} \epsilon \mathrm{d}\nu(x) + \int_{X \setminus A_i} 1 \mathrm{d}\nu(x)}\\
	&\geq
	\frac{\int_{A_i} \epsilon \mathrm{d}\nu(x)}
	{\int_{A_i} 1 \mathrm{d}\nu(x) + \int_{X \setminus A_i} 1 \mathrm{d}\nu(x)}\\
	&= \epsilon \nu(A_i)
	\end{align}
	Thus, when evaluating the modified $p$ for the learned distribution
	\begin{equation}
	p = \min_i \{\mu_k(A_i) / \nu(X_{free}) \} \geq \epsilon \min_i \{\nu(A_i) / \nu(X_{free}) \}
	\end{equation}
	One key difference between the original proof and our method is that the samples drawn are no longer independent, as the acceptance or rejection of a sample can influence future samples. However, the probability of drawing a sample from $A_i$ some $K$ number of times is lowerbounded by $(\epsilon p) ^K$ since each sample has at least $\epsilon \nu(A_i)$ probability of being drawn. Thus, the probability that the modified distribution draws $K$ successful samples from $N$ tries is lowerbounded by the probability of drawing $K$ successful independent samples out of $N$ from a biased coinflip with $p^\prime=\epsilon p$.
	
	Thus, this method simply scales the probability $p$ of the original proof by a constant factor, which does not change the proof in anyway, preserving probabilistic completeness.
	
	\section{Implementation Details}\label{ImplementationDetails}
	This section briefly describes the details of the reward function and the policy neural network so that experiments may be replicated.
	
	\begin{figure}
		\centering
		\includegraphics[width=0.9\columnwidth]{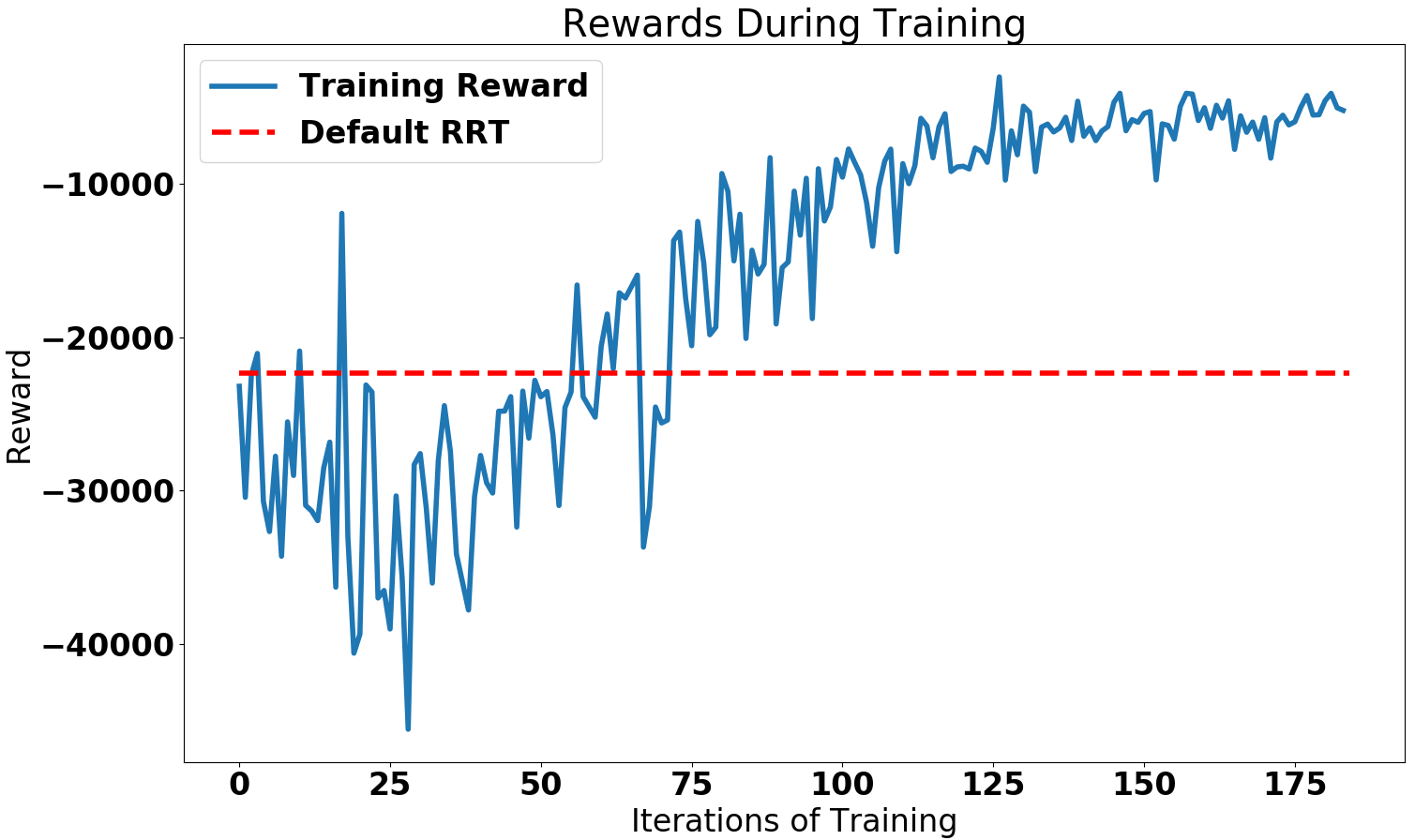}
		\caption{Rewards while training RRT.}
		\label{enva_training}
	\end{figure}
	
	\begin{figure}
		\centering
		\includegraphics[width=\columnwidth]{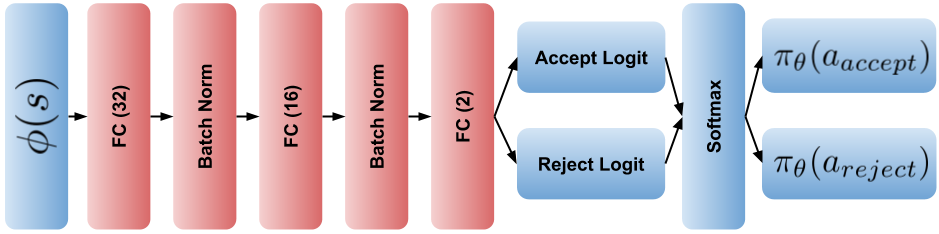}
		\caption{Policy Network Architecture. FC($N$) stands for a Fully Connected Layer with $N$ neurons.}
		\label{policynn}
	\end{figure}
	
	\subsection{Reward Function}
	The reward function $r(s, a)$ used is chosen to reflect the computation time of the planning algorithm. 
	\begin{equation}
	r(s_t, a_t) = -(\lambda_1 1 + \lambda_2 n_{node, t} + \lambda_3 n_{collision, t})
	\end{equation}
	$\lambda_1$ is a small value that represents the cost of sampling.  $n_{node, t}$ is the number of nodes added to the tree in iteration $t$ and $n_{collision, t}$ is the number of collisions checks performed in iteration $t$.
	$\lambda_2, \lambda_3$ are simply scaling factors (the experiments in this paper use $\lambda_1 = 0.01$, $\lambda_2 = \lambda_3 = 1$.). 
	Note that the total reward $\sum_{t=0}^{T}r(s_t, a_t)$ will simply be the scaled total number of nodes plus the scaled total number of collisions plus the scaled total number of samples drawn from $\nu$.
	The reward function is designed to reflect the operations that take the majority of the time: extending the tree and collision checking. The reward function can be made more elaborate, or be nonlinear, but this form is used for simplicity. In practice, this method can be made more accurate by measuring the time of each operation (collision check, node expansion, etc.) to compute the weighting factors $\lambda_i$. In addition, the rewards are normalized by their running statistics so that all problem types can have similar reward ranges.

	\begin{figure*} 
		\centering
		\begin{subfigure}{1.5\columnwidth}
			\includegraphics[width=\columnwidth]{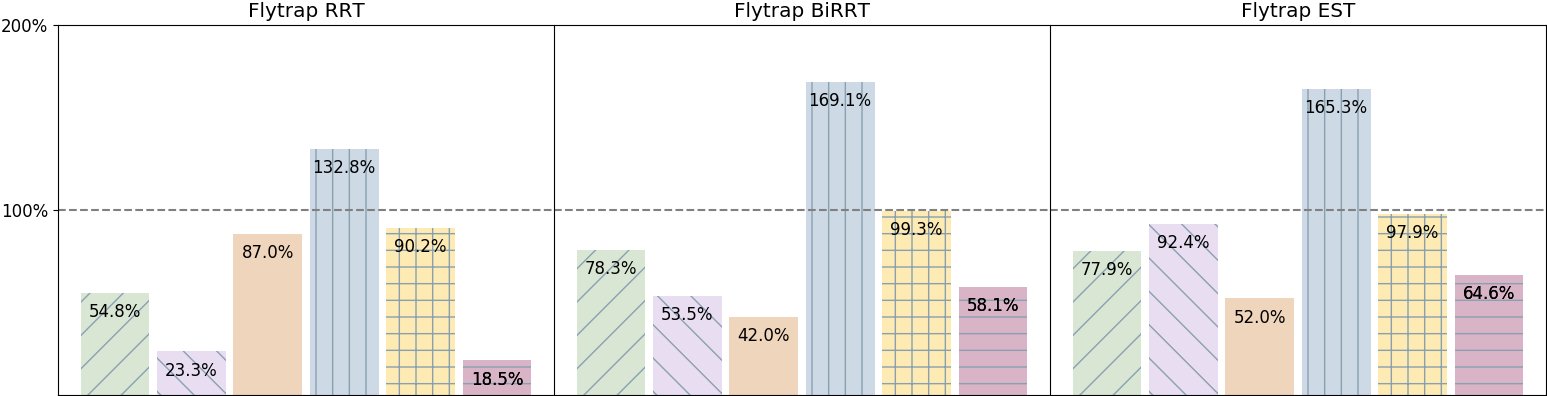}
		\end{subfigure}\hfill
		\begin{subfigure}{0.54\columnwidth}
			\includegraphics[width=\columnwidth]{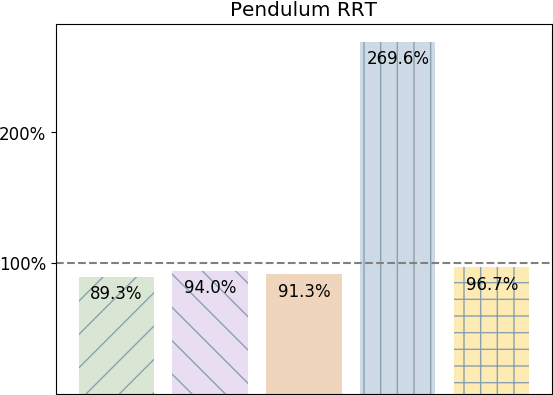}
		\end{subfigure}\hfill\\
		
		\begin{subfigure}{2\columnwidth}
			\includegraphics[width=\columnwidth]{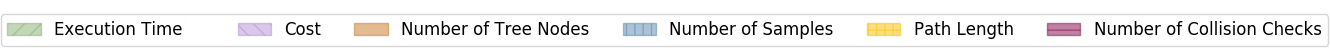}
		\end{subfigure}
		\caption{Results of 100 runs of each planner on the test environment for the Flytrap and Pendulum environments. Each bar shows the ratio of the learned planner's metric to the unmodified planner (over 100\% means more than the orginal planner).}
		\label{flytrapstats}
	\end{figure*}

	\subsection{Policy and Value Networks}
	In this work, the policy $\pi_\theta$ is a neural network that outputs probabilities of acceptance and rejection. The choice in using a neural network to represent the policy is due to the flexibility of functions they can represent. Initial results showed that a simple model like logistic regression can be insufficient in complicated environments. In addition, with neural networks, there is no need to select basis functions to introduce nonlinearities.
	
	The network used is a relatively small two layer perceptron network (the inference must be fast as this function is run many times in the inner loop of the algorithm). For reference, the network evaluted a sample in around 3.59 microseconds using only the cpu of a typical laptop. The input $\phi(s)$ is passed through two hidden layers with 32 and 16 neurons and rectified linear activation. There is a batchnorm operation \cite{Ioffe2015} after each hidden layer. The second batchnorm layer is passed to a final fully connected layer with 2 outputs that represent the logit for accepting or rejecting the sample. The logit is fed into a softmax operation to obtain the probabilities. Additionally, the logits are modified so that all probabilities lie between 0.05 and 0.95. This is so that $\pi_\theta(a_{reject} | s) < 1$ in order to guarantee that $\mu_k$ is a valid probability distribution. This also allows the policy to always have a small chance of accepting or rejecting, which is useful for exploration in the reinforcement learning algorithm. The policy network is shown in Fig. \ref{policynn}.
	
	The neural network for $V(\phi(s))$ is similar to the policy network. The only difference is that the output layer is a single neuron representing the value. All networks are trained with the Adam optimizer \cite{Kingma2014} with a learning rate of $0.001$.

	The implementation code is available at \url{https://github.com/chickensouple/learning_implicit_distributions}

	\section{Experiments}\label{Experiments}
	There are experiments done in three sets of environments. First, we test the algorithm on three different planners in a simulated FlyTrap environment. This allows us to analyze the learned policies and behavior in detail in a simplified world. Next, the algorithm is tested on a pendulum environment to analyze its performance with dynamical systems. Then, we apply the algorithm to a more complicated 7 degree of freedom robotic arm to show performance on a real system.

	\subsection{2D Flytrap}
	\begin{figure}
		\centering
		\begin{subfigure}{.3\columnwidth}
			\includegraphics[width=\columnwidth]{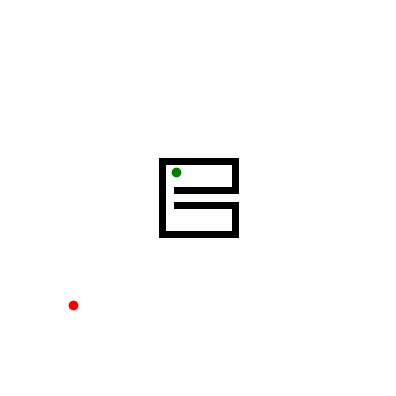}
			\caption{Train}
			\label{flytrapAtrain}
		\end{subfigure}\hfill
		\begin{subfigure}{.3\columnwidth}
			\includegraphics[width=\columnwidth]{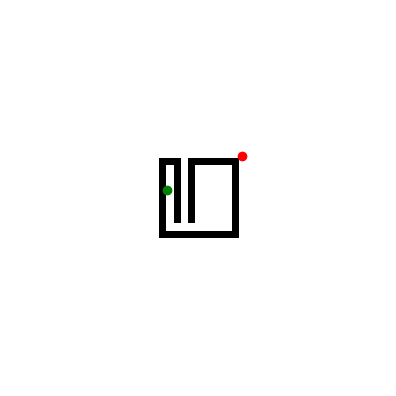}
			\caption{Test}
			\label{flytrapAtest}
		\end{subfigure}\hfill
		\begin{subfigure}{.3\columnwidth}
			\includegraphics[width=\columnwidth]{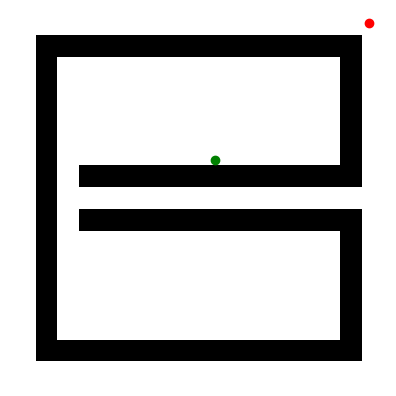}
			\caption{BallTree}
			\label{flytrap_b}
		\end{subfigure}\hfill
		
		\caption{Various Flytrap environments. The green dot is an example starting location and the red dot is an example goal location.}
		\label{flytrapA}
	\end{figure}
	
	\begin{figure*}
		\centering
		\begin{subfigure}{.6\columnwidth}
			\includegraphics[width=\columnwidth]{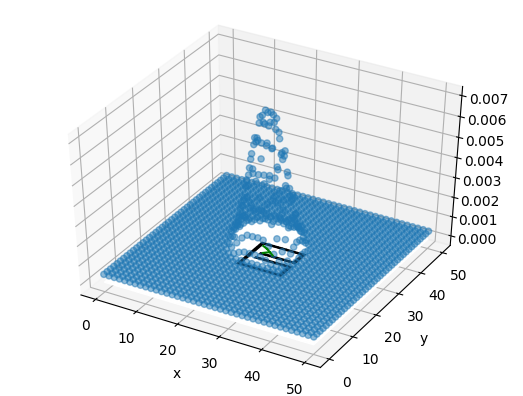}
		\end{subfigure}\hfill
		\begin{subfigure}{.6\columnwidth}
			\includegraphics[width=\columnwidth]{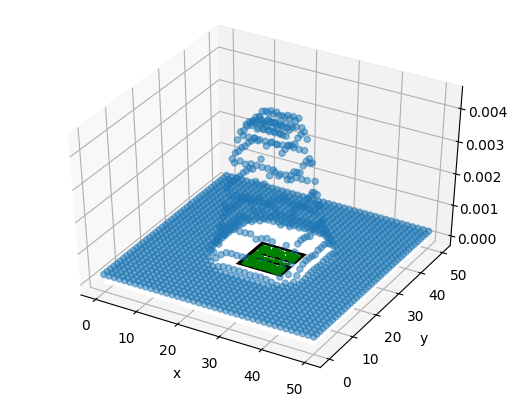}
		\end{subfigure}\hfill
		\begin{subfigure}{.6\columnwidth}
			\includegraphics[width=\columnwidth]{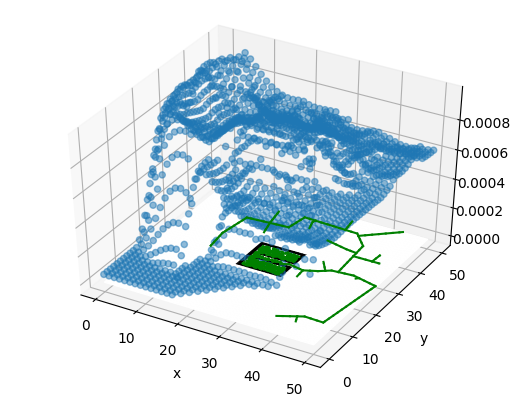}
		\end{subfigure}\hfill
		\caption{Learned probability distributions for RRT. While the policy is the same, the distributions change as the RRT search progresses. For each figure, the bottom plane shows the environment with a green search tree, while the blue dots show sampled points representing the learned distribution.}
		\label{Adistributions}
	\end{figure*}
	
	The first experiment run is that of the 2D Flytrap. This environment is used as a benchmark in \cite{Yershova2005} and \cite{Shkolnik2011} as an example of a hard planning problem. It is difficult to solve because of the thin tunnel that must be sampled in order to find a path to the goal. The training and testing environments are shown in Fig. \ref{flytrapA}. Three different planners are tested on the environment: RRT with Connect function \cite{Kuffner2000}, Bidirectional RRT (BiRRT) \cite{Kuffner2000}, and EST. An example of the training curve is shown in Fig. \ref{enva_training}. 
	
	For RRT, the feature used is the distance to the nearest tree node minus the distance of that tree node to its nearest obstacle. For BiRRT, the feature used is the distance to the current tree being expanded minus the distance of that tree node to its nearest obstacle. For EST, there are a few choices for how to modify the sampling. In this experiment, we chose to modify the probability of picking nodes in the tree for expansion (the alternative being modifying the probability of how to pick nodes to expand to) since the choice of node has a larger effect on the algorithm's performance. The features used are two dimensional: the nearest obstacle to the node, as well as the number of nodes in a certain radius (this is the same as $w(x)$ used in the original EST paper \cite{Hsu1997}). 
	
	For each planner, the original policy of always accepting samples is compared against the policy trained on the environment shown in Fig. \ref{flytrapAtrain}. The results in Fig. \ref{flytrapstats} show the statistics over 100 run. The average of each metric tracked for the planners is compared.
	
	For all planners, the number of collision checks is reduced while the number of samples drawn is increased. In RRT, it is reduced around five times. The tradeoff between collision checks and number of samples saves overall execution time. In addition, the decreasing the tree size and reducing collision checks does not decrease the quality of the paths found. For each planner, the path found by the trained policy is equivalent in length or sometimes shorter, despite not explicitly optimizing for path length.
	
\begin{figure}
		\centering
		\includegraphics[width=\columnwidth]{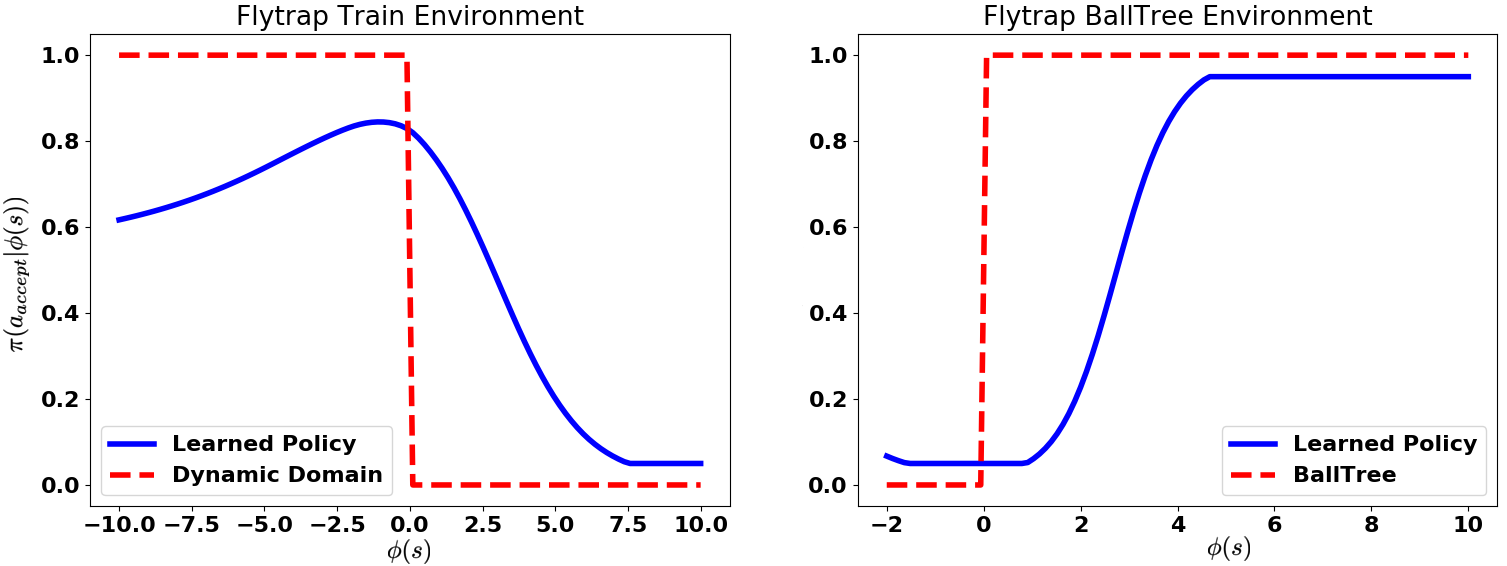}
		\caption{Comparison between learned policies and BallTree and Dynamic Domain RRT.}
		\label{policy_comparison}
	\end{figure}

	Next, the policies learned for RRT are analyzed. The learned policy rejects samples that are far away from the tree with higher probability. This is similar to the strategy that is suggest by Dynamic Domain RRT \cite{Yershova2005}, in which the ideal version of it rejects all samples that are further away from the tree than the closest obstacle. However, for Flytrap environments where the space outside of the Flytrap is not a large fraction of the space, the strategy suggested by BallTree \cite{Shkolnik2011} is more effective. BallTree rejects all samples that are closer to the tree than the nearest obstacle. It is curious that for very similar types of environments, the policies that work better for each are almost complete opposites! This shows a need to use the data itself to tune a rejection sampling policy. When training on the different sized environment shown in Fig. \ref{flytrap_b}, the policies learned to exhibit behaviour similar to BallTree. The policy trained in the larger environment rejects samples further from the tree, and the policy trained in the smaller environment rejects samples that are closer to the tree as shown in Fig. \ref{policy_comparison}. The distributions encountered during the search process are visualized in Fig. \ref{Adistributions} by sampling a uniform grid in the statespace and using Eq. \ref{rejectionsampling} to compute discretized probabilities for sampling each point.

	\subsection{Pendulum Task}
	In addition to the flytrap environment, experiments were done on a planar pendulum to test the effectiveness of it on a dynamical system. The pendulum starts at the bottom and needs to reach the top. It is control limited so it must plan a path that increases its energy until it can swing up. In this experiment, we used a steering function that randomly samples control actions and time durations. This is a common steering function that may be used in more complicated systems \cite{li2016asymptotically}. The results are shown in Fig. \ref{flytrapstats}. Number of collision checks is not included as for this particular experiments as there are no obstacles to collide with. The features used are 1) the difference between the goal angle and the current angle and 2) the difference in angular velocities. The policy learns to reject samples that are not likely to lead to the goal state, which saves the execution time otherwise spent computing the steering function.
	
	\subsection{7 Degree of Freedom Arm}
\begin{figure}
		\centering
		\includegraphics[width=0.8\columnwidth]{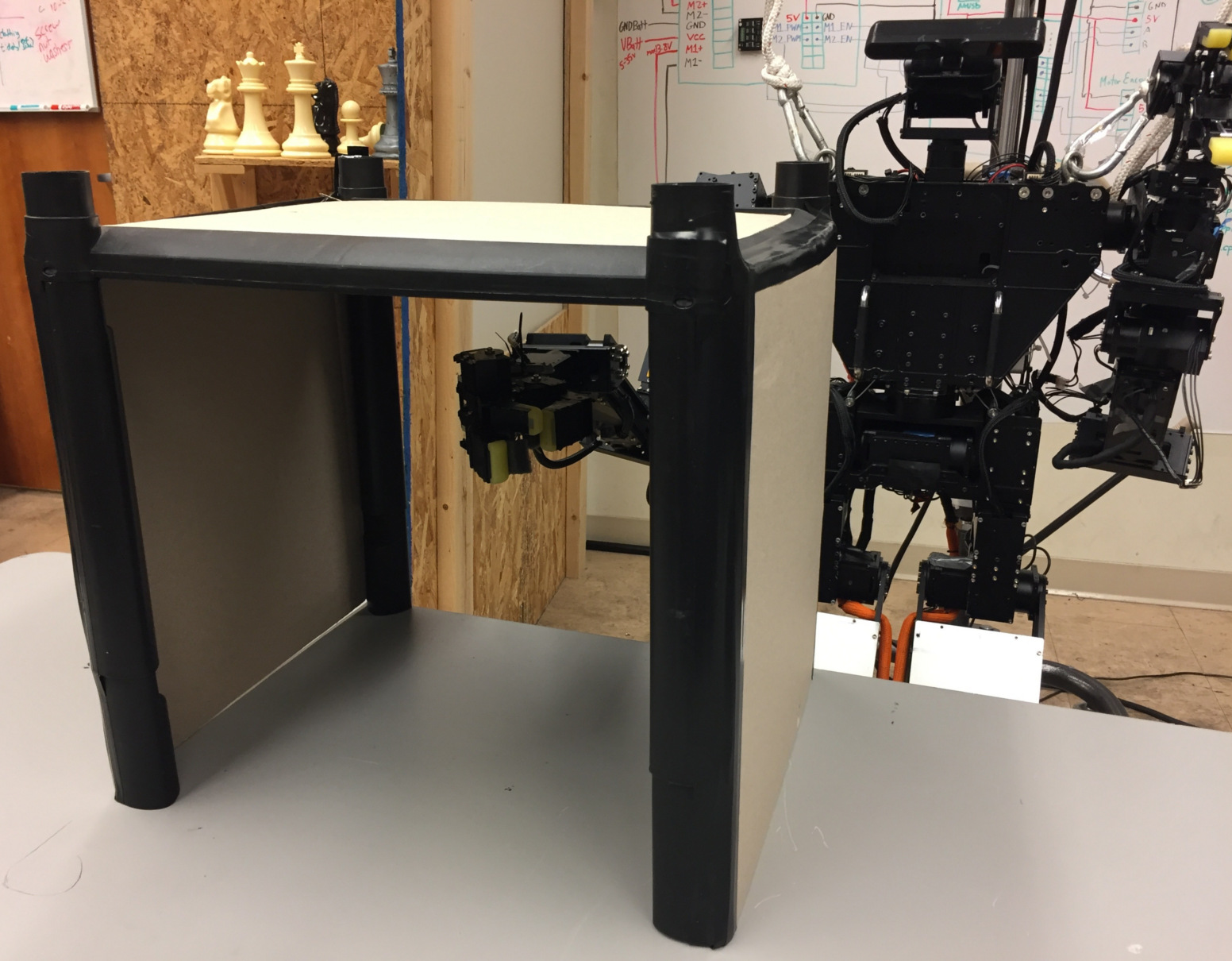}
		\caption{The Thor robot in a test of the tabletop environment.}
		\label{thor}
	\end{figure}

	The algorithm is also tested on the 7 degree of freedom arm of the Thor robot (Fig. \ref{thor}). This experiment is used to validate the method in a higher dimensional space and in a realistic environment. Thor is given tasks to move its arm to various difficult to reach places in assorted tabletop environments. The environments consists of crevices for Thor to reach into and obstacles to block passages. The base planner used is BiRRT, with a four dimensional feature space (EST and RRT were not used as the planning took too long). The first three features are the distances of various joints to the closest obstacle, and the last feature is the distance of the current configuration to the goal. Two very different environments were used for training, and a third environment distinct from the first two was used for testing.
	
	\begin{figure}
		\centering
		
		\begin{subfigure}{0.8\columnwidth}
			\includegraphics[width=\columnwidth]{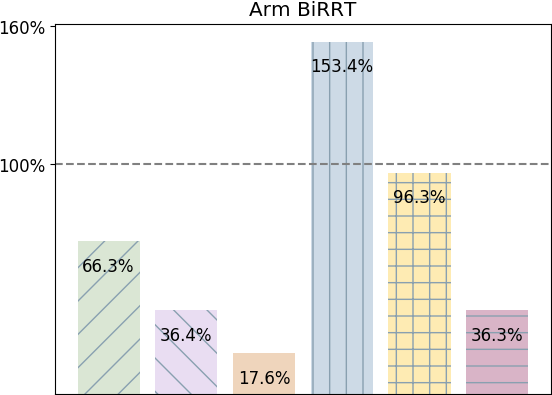}
		\end{subfigure}\hfill
		\begin{subfigure}{0.8\columnwidth}
			\includegraphics[width=\columnwidth]{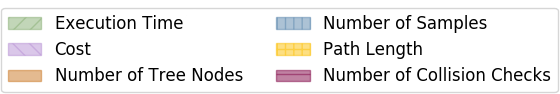}
		\end{subfigure}\hfill
		
		\caption{Comparison of results of BiRRT in tabletop environments. Each bar shows the ratio of the learned planner's metric to the unmodified planner}
		\label{arm_stats}
	\end{figure}

	The results of the arm experiments are shown in Fig. \ref{arm_stats}. The figure details the statistics over successful plans over 100 runs of the planner. Our algorithm had 97\% success rate in finding a path, while the original had 96\% when the number of samples drawn is limited to 100,000 (this difference is too small to make any claims). Similar to the Flytrap experiments, a policy is learned that trades off extra samples for a vastly reduced number of collision checks and nodes in the tree. On the test environment, the number of nodes in the tree is more than 5 times less and uses 2.7 times less collision checks. In addition, the variance of the results is greatly reduced when using the learned distribution.
	
	Next, the policies learned for the Thor arm are examined to see what aspect of the environment it is exploiting. We note that the probability increases as 1) the distance of the configuration to the goal is lower, or 2) the workspace distance of the later joints is closer to an obstacle. This policy makes a lot of intuitive sense. Samples are concentrated near the surface of the table and objects, probing the surface for a good configuration.

	%\addtolength{\textheight}{-0.2cm}   % This command serves to balance the column lengths
	% on the last page of the document manually. It shortens
	% the textheight of the last page by a suitable amount.
	% This command does not take effect until the next page
	% so it should come on the page before the last. Make
	% sure that you do not shorten the textheight too much.
    
	\section{CONCLUSIONS}\label{Conclusion}
	Sampling distributions in sampling-based motion planners are a vital component of the algorithm that affects how many times computationally expensive subroutines such as collision checks are run. While the method presented can improve planning times by modifying the sampling distribution, it is not the whole solution for all problem types. In maps where the thin tunnel issue is more pronounced, rejection sampling does not alleviate the main dilemma of how to sample the thin tunnel. However, this method can be easily combined with existing techniques such as \cite{Zhang2008, Yang2004, Choudhury2016} to improve performance. 
	
	In addition, this paper does not directly address the problem of finding the optimal solution. The authors believe that an offline method for generating sampling distributions is not the solution for that aspect of planning. Instead, this method can be applied to existing optimal planners. It can, for instance, be used to find the first solution faster in Informed RRT* \cite{Gammell2014} or BIT* \cite{Gammell2015} to improve upon existing methods. 
	
	In conclusion, this paper presents a general way to obtain good rejection sampling schemes for sampling-based motion planners. The process can be seen as a way of encoding the prior knowledge of the environments into the rejection policy by learning from previous searches in similar environments and is shown to be effective in practice. 
	
	\section*{ACKNOWLEDGMENT}
	
	This material is based upon work supported by the National Science Foundation Graduate
	Research Fellowship Program under Grant No. DGE-1321851. Any opinions,
	findings, and conclusions or recommendations expressed in this material are those of the
	authors and do not necessarily reflect the views of the National Science Foundation. The authors would like to thank Bhoram Lee and Min Wen for thoughtful conversations about the paper.
	
	\bibliographystyle{ieeetr}
	\bibliography{bib}

	%%%%%%%%%%%%%%%%%%%%%%%%%%%%%%%%%%%%%%%%%%%%%%%%%%%%%%%%%%%%%%%%%%%%%%%%%%%%%%%%

	%%%%%%%%%%%%%%%%%%%%%%%%%%%%%%%%%%%%%%%%%%%%%%%%%%%%%%%%%%%%%%%%%%%%%%%%%%%%%%%%

	%%%%%%%%%%%%%%%%%%%%%%%%%%%%%%%%%%%%%%%%%%%%%%%%%%%%%%%%%%%%%%%%%%%%%%%%%%%%%%%%

\end{document}